\documentclass[10pt,twocolumn,letterpaper]{article}

\usepackage{wacv}
\usepackage{times}
\usepackage{epsfig}
\usepackage{graphicx}
\usepackage{amsmath}
\usepackage{amssymb}

\usepackage{amsmath,amssymb} 
\usepackage{color}
\usepackage{cite}

\usepackage{subcaption}
\usepackage{adjustbox}
\usepackage{floatrow}
\def\wacvPaperID{250} 

\wacvfinalcopy 

\ifwacvfinal
\def\assignedStartPage{1} 
\fi
\ifwacvfinal
\usepackage[breaklinks=true,bookmarks=false]{hyperref}
\else
\usepackage[pagebackref=true,breaklinks=true,colorlinks,bookmarks=false]{hyperref}
\fi

\ifwacvfinal
\else
\fi

\begin{document}

\title{Real-Time Uncertainty Estimation in Computer Vision via Uncertainty-Aware Distribution Distillation}

\makeatletter
\def\and{%
  \end{tabular}%
  \hskip 0.1em \@plus.17fil\relax
  \begin{tabular}[t]{c}}
\makeatother

\author{Yichen Shen \thanks{The first two authors contributed equally to this work.}\\
Samsung Inc, USA\\
\and
Zhilu Zhang \footnotemark[1]\\
Cornell University\\
\and
Mert R.~Sabuncu\\
Cornell Univerisity\\
\and
Lin Sun\\
Samsung Inc, USA\\
}

\maketitle

\begin{abstract}
Calibrated estimates of uncertainty are critical for many real-world computer vision applications of deep learning.
While there are several widely-used uncertainty estimation methods, dropout inference\cite{gal2016dropout} stands out for its simplicity and efficacy. 
This technique, however, requires multiple forward passes through the network during inference and therefore can be too resource-intensive to be deployed in real-time applications. 
We propose a simple, easy-to-optimize distillation method for learning the conditional predictive distribution of a pre-trained dropout model for fast, sample-free uncertainty estimation in computer vision tasks. 
We empirically test the effectiveness of the proposed method on both semantic segmentation and depth estimation tasks, 
and demonstrate our method can significantly reduce the inference time, enabling real-time uncertainty quantification, while achieving improved quality of both the uncertainty estimates and predictive performance over the regular dropout model.
\end{abstract}

\section{Introduction}
Uncertainty exists in many machine learning problems due to noise in the observations and incomplete coverage of domain. How certain can we trust the model built upon limited yet imperfect data? Reliable uncertainty estimates are crucial for trustworthy applications such as medical diagnosis and autonomous driving. Many algorithms have been proposed to estimate the uncertainty of neural networks (NN)\cite{kingma2015variational, louizos2017multiplicative,blundell2015weight,wu2018deterministic}. Among these, the MC dropout\cite{gal2016dropout} is arguably one of the most popular approaches due to its simplicity and scalability. This approach has been adopted in different computer vision tasks recently\cite{feng2018towards,bertoni2019monoloco}. A follow-up work~\cite{kendall2017uncertainties}  further enhanced quality of uncertainty estimates by incorporating both aleatoric and epistemic uncertainty into deep neural networks. 
\begin{figure}
\centering
\includegraphics[width=1.0\linewidth]{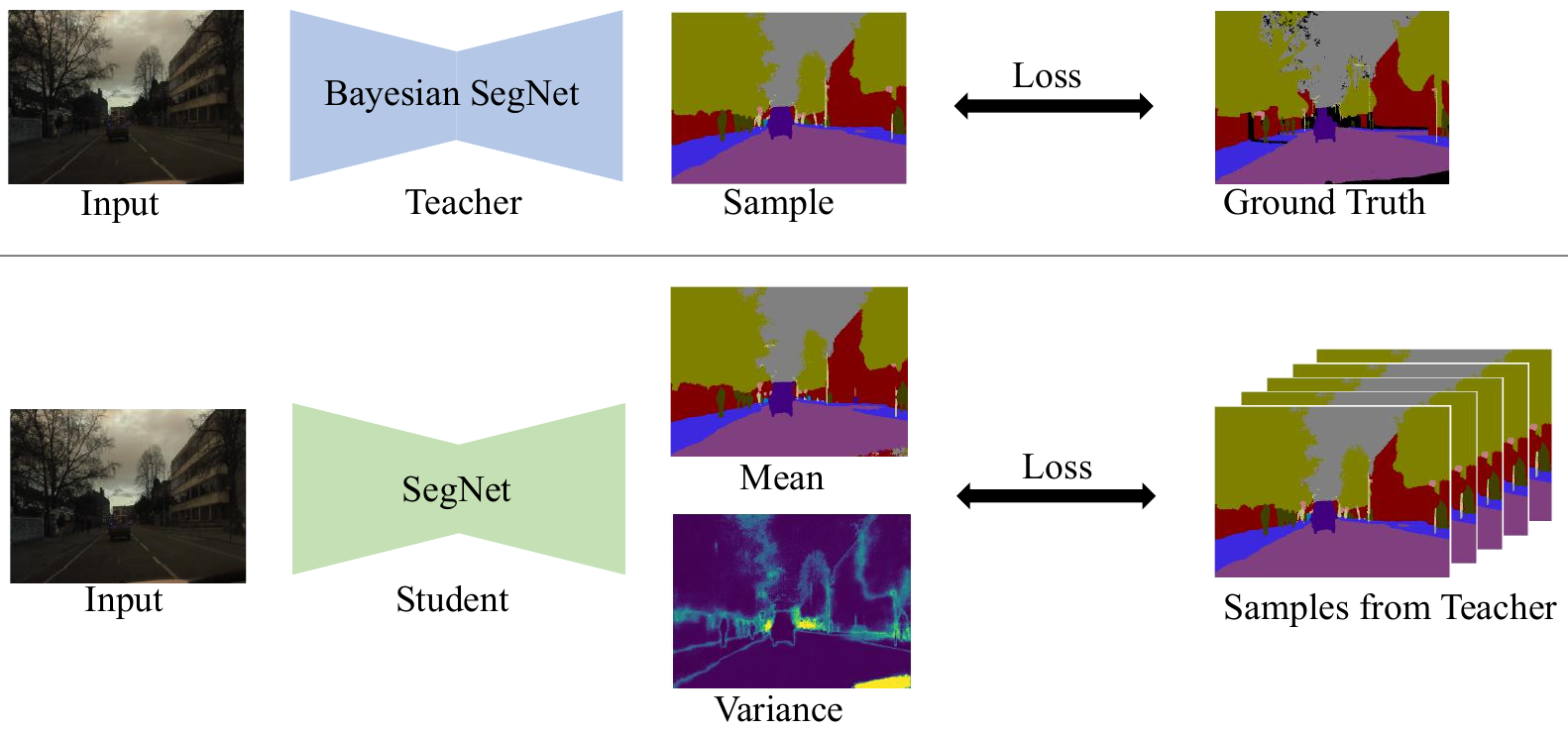}
\caption{An illustration of the proposed method. Given a trained teacher, a deterministic student is used to approximately parameterize the predictive distribution of the teacher model, enabling sample-free uncertainty estimation.}
\label{Illustration_figure}
\end{figure}
Despite its success, MC dropout requires test-time sampling to obtain uncertainty. This costly sampling process can introduce severe latency in real-time prediction tasks such as the perception system of self-driving vehicles and lead to undesired consequences. 

In order to eliminate the expensive dropout sampling at the test-time, prior work \cite{bulo2016dropout} has explored distilling knowledge from MC dropout samples of a teacher model into a student network (Dropout Distillation or DD).
Nevertheless, DD has several limitations. Specifically, the student model only learns from the predictive means of the dropout teacher model, and the dispersion of the teacher's prediction which entails important uncertainty information associated with the predictions~\cite{malinin2018predictive}, is completely neglected in their approach. 
To address the problem, in this paper, we propose an easy-to-optimize, generally applicable distillation framework for fast, sample-free uncertainty estimation. Specifically, we approximate the entire predictive distribution produced by a MC-dropout teacher with flexible parametric distributions. At test time, the parameters of the distribution are output by a single deterministic student network to obtain reliable uncertainty estimates in one forward pass. In addition, we show that our method can distill both epistemic and aleatoric uncertainty with little extra computation.

We examine the effectiveness of the proposed method on regression and classification with high resolution, real-world datasets. For regression, we experiment on monocular depth estimation using NYU Depth V2\cite{SilbermanECCV12} and KITTI\cite{Geiger2013IJRR}. For classification, we experiment on semantic segmentation using CamVid\cite{BrostowSFCECCV08} and VOC2012\cite{pascal-voc-2012}. In addition to significant faster inference time, quantitative and qualitative results show the student network produces uncertainty estimates of better quality than those of the teacher model, i.e. MC dropout pre-trained model. We also demonstrate the predictive mean and uncertainty obtained with our method are superior to those learned from DD\cite{bulo2016dropout}. 


\section{Related Work}

Uncertainty estimation can be obtained for deep learning in a principled manner through Bayesian neural networks\cite{mackay1992practical, neal2012bayesian}. However, they typically suffer from significant computational burdens due to the intractability of posteriors. As such, computational tractability has been a primary focus of research. One such direction is through Markov Chain Monte Carlo (MCMC)\cite{neal2012bayesian}. For instance, stochastic gradient versions of MCMC have been proposed\cite{chen2014stochastic, welling2011bayesian, ma2015complete, gong2018meta} to scale MCMC method to large datasets. Nevertheless, these approaches can be difficult to scale to high-dimensional data. An alternative solution is through variational inference in which parametric distributions are used to approximate the intractable true posteriors of the weights of neural networks\cite{kingma2015variational,louizos2017multiplicative,blundell2015weight,wu2018deterministic}. However, the variational inference can suffer from sub-optimal performances\cite{blier2018description}.

There have also been Non-Bayesian techniques for uncertainty estimation. For instance, an ensemble of randomly-initialized NNs\cite{lakshminarayanan2017simple} has shown to be effective. However, it requires training and saving multiple NNs, which can be costly in practice.
Methods to more efficiently obtaining ensembles exist\cite{geifman2018bias,huang2017snapshot}, but these can come at the cost of quality of uncertainty estimates. 
Most of the above-mentioned methods require multiple forward passes of NNs at test time, which prohibits their deployment in real-time computer vision systems. Several techniques have been proposed to speed up uncertainty estimation. Postels et al.\cite{postels2019sampling} proposed a method for sampling-free uncertainty estimation through variance propagation. However, the simplistic assumptions they made about the covariance matrices of NN activation might lead to inaccurate approximations. Another approach speeds up the sampling process by leveraging the temporal information in videos\cite{huang2018efficient}. However, the method cannot be generically applied and results in a non-trivial drop in predictive performance. Ilg et al.\cite{ilg2018uncertainty} proposed the use of a multi-hypotheses NN with a novel loss function to obtain sample-free uncertainty estimates in the optical flow estimation task. However \cite{ilg2018uncertainty} uses an additional network to merge the hypotheses that incurs extra memory cost at inference time.

Distillation-based methods have also been explored. 
\cite{bulo2016dropout} proposed to distill predictive means from a dropout teacher to a student network. As addressed above, this leads to the loss of the epistemic uncertainty of the MC dropout teacher. Most similar to our method is \cite{malinin2019ensemble}, which uses the Dirichlet distribution to approximate predictive distribution of an ensemble of networks. However, the proposed method requires a large size of ensemble to effectively train a student network, which can be prohibitively expensive for challenging computer vision tasks. In addition, the proposed use of the Dirichlet distribution is not only practically hard to optimize, but also not applicable for regression tasks. 

\section{Method}
Suppose we have a dataset $\mathcal{D} = (\boldsymbol{X}, \boldsymbol{Y}) = \{ (\boldsymbol{x}_i, \boldsymbol{y}_i) \}_{i = 1}^n$, where each $ (\boldsymbol{x}_i, \boldsymbol{y}_i) \in (\mathcal{X} \times \mathcal{Y}) $ is i.i.d. and $ \mathcal{X} \subseteq \mathbb{R}^d$ corresponds to the feature space. Most tasks in computer vision can be considered as either regression or classification. For regression, $ \mathcal{Y} \subseteq \mathbb{R}^k$ for some integer $k$, and in the context of $k$-class classification, $\mathcal{Y} = \{ 1, \cdots , k \}$ is the label space. We define $f_{\boldsymbol{w}}(\boldsymbol{x})$ to be a neural network such that $f: \mathcal{X} \rightarrow \mathcal{Y} $, and $\boldsymbol{w} = \{ W_i \}_{i = 1}^L$ corresponds to the parameters of the network with L-layers, where each $W_i$ is the weight matrix in the i-th layer. We define the model likelihood $p(\boldsymbol{y}|\boldsymbol{x}, \boldsymbol{w} ) = p(\boldsymbol{y} | f_{\boldsymbol{w}}(\boldsymbol{x}))$.
For regression tasks, it is common to assume $p(\boldsymbol{y} | f_{\boldsymbol{w}}(\boldsymbol{x})) = \mathcal{N}(f_{\boldsymbol{w}}(\boldsymbol{x}), \sigma^2)$, for some noise term $\sigma$. For classification tasks, $p(\boldsymbol{y} | f_{\boldsymbol{w}}(\boldsymbol{x})) = \text{Softmax}(f_{\boldsymbol{w}}(\boldsymbol{x}))$ is commonly assumed.
To capture epistemic uncertainty, we put a prior distribution on the weights of the network, $p(\boldsymbol{w})$. 
A common choice is the zero mean Gaussian $\mathcal{N}(0,I)$. Bayes Theorem can then be used to obtain the posterior $p(\boldsymbol{w} |\boldsymbol{X}, \boldsymbol{Y}) = p(\boldsymbol{Y}|\boldsymbol{X}, \boldsymbol{w})p(\boldsymbol{w}) / p(\boldsymbol{Y}|\boldsymbol{X})$, with which the predictive distribution can be determined by
\begin{align}
p(\boldsymbol{y}| \boldsymbol{x}, \mathcal{D_{\text{train}}}) = \int p(\boldsymbol{y}|\boldsymbol{x}, \boldsymbol{w})p(\boldsymbol{w}|\mathcal{D_{\text{train}}})  \, d \boldsymbol{w}.
\end{align} 

\subsection{Preliminary: Dropout for Bayesian Deep Learning}
The marginal distribution $p(\boldsymbol{Y}|\boldsymbol{X})$, and thus $p(\boldsymbol{w} |\boldsymbol{X}, \boldsymbol{Y})$ are often intractable. 
Variational inference uses a tractable family of distributions $q_{\theta}(\boldsymbol{w})$ paramaterized by ${\theta}$ to approximate the true posterior $p(\boldsymbol{w} |\boldsymbol{X}, \boldsymbol{Y})$, thereby turning the problem into a tractable optimization task. MC dropout, which casts the dropout regularization as approximate Bayesian inference, is one such example\cite{gal2016dropout}. It involves training NNs with dropout after each weight layer. 
With an optimized model, the approximate predictive distribution is given by
$q(\boldsymbol{y} | \boldsymbol{x}, \mathcal{D_{\text{train}}}) = \int p(\boldsymbol{y}|\boldsymbol{x}, \boldsymbol{w})q_{\theta}(\boldsymbol{w}) \, d \boldsymbol{w}$.
The integral can be approximated through performing Monte Carlo integration over $q_{\theta}(\boldsymbol{w})$. This corresponds to dropout at test time. In classification for example, 
\begin{align}
p(y = c | \boldsymbol{x}, \mathcal{D_{\text{train}}}) \approx \frac{1}{T} \sum _{t = 1}^T  \text{Softmax}(f_{\boldsymbol{w}_t}(\boldsymbol{x})),
\end{align}
where $\boldsymbol{w}_t \sim q_{\theta}(\boldsymbol{w})$ are dropout samples from the NN.

Epistemic uncertainty can be computed with the approximate inference framework as derived above. For regression, epistemic uncertainty is captured by the predictive variance, which can be approximated by computing the variance of the dropout approximate distribution:
\begin{align}
    \boldsymbol{\sigma}_{\boldsymbol{y}}^2 \approx \frac{1}{T} \sum _{t = 1}^T  f_{\boldsymbol{w}_t}(\boldsymbol{x})^T f_{\boldsymbol{w}_t}(\boldsymbol{x}) - \boldsymbol{\mu}_{\boldsymbol{y}}^T \boldsymbol{\mu}_{\boldsymbol{y}},
\end{align}
where $\boldsymbol{\mu}_{\boldsymbol{y}} = \sum _{t = 1}^T  f_{\boldsymbol{w}_t}(\boldsymbol{x})$. In the context of classification, numerous measures have been proposed as uncertainty estimates\cite{gal2017deep}. In this paper, we use the mutual information between the predictions and the model posterior (BALD), 
\begin{align}
    \mathbb{I}\left[ \boldsymbol{y}, \boldsymbol{w} | \boldsymbol{x}, \mathcal{D}_{train} \right] = \mathbb{H}\left[ \boldsymbol{y}| \boldsymbol{x}, \mathcal{D}_{train} \right] - \mathbb{E}\left[ \mathbb{H}\left[ \boldsymbol{y}| \boldsymbol{x}, \boldsymbol{w} \right] \right]
\end{align}
as the uncertainty estimates in classification tasks. 

As proposed by Kendall and Gal\cite{kendall2017uncertainties}, aleatoric uncertainty can also be incorporated into the dropout model for concurrent estimation of both the epistemic and aleatoric uncertainty. 
To do so, an input-dependent observation noise parameter $\hat{\boldsymbol{\sigma}}^2$ is output together with the prediction
\begin{align}
    \left[ \hat{\boldsymbol{\mu}}, \hat{\boldsymbol{\sigma}}^2 \right] = f_{\boldsymbol{w}}(\boldsymbol{x}),\label{eq10}
\end{align}
where $\hat{\boldsymbol{\sigma}}^2$ is a vector with the same dimension as $\hat{\boldsymbol{y}}$ that represents a diagonal covariance matrix. $\hat{\boldsymbol{\sigma}}$ can be optimized with maximum likelihood estimation by assuming that aleatoric uncertainty follows a parametric distribution (e.g Gaussian) for regression. For classification, this Gaussian distribution is placed over the logit space. Although the epistemic and aleatoric uncertainties are not mutually exclusive, the total uncertainty can be approximated using
\begin{align}
    \textbf{Var}(\boldsymbol{y}) \approx \frac{1}{T} \sum _{t = 1}^T  \hat{\boldsymbol{\mu}}_t^T \hat{\boldsymbol{\mu}}_t - \boldsymbol{\mu}_{\boldsymbol{y}}^T \boldsymbol{\mu}_{\boldsymbol{y}} + \frac{1}{T} \sum _{t = 1}^T \hat{\boldsymbol{\sigma}}_t^2\label{eq11}
\end{align}
where $\boldsymbol{\mu}_{\boldsymbol{y}} = \sum _{t = 1}^T  \hat{\boldsymbol{\mu}}_t$ and the first and second term in the above expression approximates the epistemic and aleatoric uncertainties respectively.

\subsection{A Teacher-Student Paradigm for Sample-free Uncertainty Estimation}
Despite the success of the MC dropout, inferring uncertainty at test-time often requires multiple forward passes to generate samples of prediction, limiting its application to many time-sensitive applications. In this paper, we propose to use a deterministic neural network $f_{\boldsymbol{\phi}}(\boldsymbol{x})$ to parameterize a distribution $r(\boldsymbol{y} | \boldsymbol{x}, \mathcal{D_{\text{train}}})$ that approximates the predictive distribution $q(\boldsymbol{y} | \boldsymbol{x}, \mathcal{D_{\text{train}}})$ of the dropout model. Specifically, $f_{\boldsymbol{\phi}}(\boldsymbol{x})$ learns to directly output the parameters of $r(\boldsymbol{y} | \boldsymbol{x}, \mathcal{D_{\text{train}}})$. When trained, $f_{\boldsymbol{\phi}}(\boldsymbol{x})$ only requires one forward pass to infer both predictive mean and uncertainty from the parameterized distribution $r(\boldsymbol{y} | \boldsymbol{x}, \mathcal{D_{\text{train}}})$, thus eliminating expensive sampling processes at test time.

Training $f_{\boldsymbol{\phi}}(\boldsymbol{x})$ is straight-forward using a teacher-student paradigm similar to the knowledge distillation\cite{hinton2015distilling}. We first train a Bayesian neural network (BNN) $f_{\boldsymbol{w}}(\boldsymbol{x})$ (e.g. a dropout model) on $\mathcal{D_{\text{train}}}$. We then generate samples of predictions from the pre-trained $f_{\boldsymbol{w}}(\boldsymbol{x})$. These samples serve as ``observations'' from the distribution $q(\boldsymbol{y} | \boldsymbol{x}, \mathcal{D_{\text{train}}})$ for $f_{\boldsymbol{\phi}}(\boldsymbol{x})$ to learn the parameters of $r(\boldsymbol{y} | \boldsymbol{x}, \mathcal{D_{\text{train}}})$ given each input $x \in \mathcal{D_{\text{train}}}$. Eventually $f_{\boldsymbol{\phi}}(\boldsymbol{x})$ learns an efficient mapping from input images to the parameters of the distribution  $r(\boldsymbol{y} | \boldsymbol{x}, \mathcal{D_{\text{train}}})$ that accurately approximates $q(\boldsymbol{y} | \boldsymbol{x}, \mathcal{D_{\text{train}}})$. For simplicity, in the following illustration we term the BNN $f_{\boldsymbol{w}}(\boldsymbol{x})$ as the teacher model and $f_{\boldsymbol{\phi}}(\boldsymbol{x})$ as the student model. 

\subsubsection{Sampling from the Bayesian Teacher}
As mentioned above, predictive samples $\{ \hat{\boldsymbol{y}}_t = f_{\boldsymbol{w}_t}(\boldsymbol{x}) \}_{t=1}^m$ are generated from the teacher to train the student. In the more complicated scenario where aleatoric uncertainty is modeled by teacher, 
we incorporate aleatoric uncertainty into each predictive sample with
\begin{align}
    \hat{\boldsymbol{y}}_t = \hat{\boldsymbol{\mu}}_t + \hat{\boldsymbol{\sigma}}_t \boldsymbol{\epsilon}, \quad \boldsymbol{\epsilon} \sim \mathcal{N}(0, I).\label{eq12}
\end{align}
where $\hat{\boldsymbol{\sigma}}_t$ is the aleatoric uncertainty output by the teacher given an input. In practice, $\hat{\boldsymbol{\sigma}}_t$ can be noisy. To stabilize training, instead of $\hat{\boldsymbol{\sigma}}_t^2$, we first compute empirical mean $\tilde{\boldsymbol{\sigma}}^2 \triangleq \frac{1}{T} \sum _{t = 1}^T \hat{\boldsymbol{\sigma}}_t^2$, and use $\tilde{\boldsymbol{\sigma}}^2$ to generate all samples $\{ \hat{\boldsymbol{y}}_t\}_{t=1}^m$. 

The larger the number of samples, the more accurate the student approximation can be to the teacher predictive distribution. However, sampling a large number of samples requires intensive computational resources. To cope with this challenge, we generate a small number of $m$ predictive samples from the teacher for each input on-the-fly at each epoch during training. 
In order to learn aleatoric uncertainty, we further get $k$ random samples from $\mathcal{N}(0, I)$ for each predictive sample $\left[ \hat{\boldsymbol{\mu}}_t, \tilde{\boldsymbol{\sigma}}^2 \right]$ (see Eq. \ref{eq12}). In practice we use $m=5$ and $k=10$. As we demonstrate in the experimental section, a small number of $m$ and $k$ per input is sufficient to learn student model with excellent performance. 

\subsubsection{Optimizing the Student} 
We use maximum likelihood estimation (MLE) to optimize $f_{\boldsymbol{\phi}}(\boldsymbol{x})$. Given the samples $\{ \hat{\boldsymbol{y}}_t\}_{t=1}^m$ generated by the teacher, we minimize the negative log likelihood for each input $\boldsymbol{x}$
\begin{align}
    \mathcal{L}_{s} = -\sum_t \log r \left( \hat{\boldsymbol{y}}_t | \boldsymbol{x} ; \boldsymbol{\phi} \right ).
    \label{mle}
\end{align}
where $r ( \hat{\boldsymbol{y}}_t | \boldsymbol{x} ; \boldsymbol{\phi})$ is parameterized by $f_{\boldsymbol{\phi}}(\boldsymbol{x})$. In order to avoid division by zero and enable unconstrained optimization of the variance, we use log variance $\boldsymbol{s} = \log(\hat{\boldsymbol{\sigma}}^{2})$ as the output of the student. Thus, we have $\left[ \hat{\boldsymbol{\mu}}, \boldsymbol{s} \right] = f_{\boldsymbol{\phi}}(\boldsymbol{x}).$

For regression problems, we use the Laplace distribution to approximate the variational predictive distribution. For simplicity, we assume independence among all the dimensions of outputs so that log variance $\boldsymbol{s}$ is a vector of the same dimension as $\hat{\boldsymbol{\mu}}$. Given the Laplace assumption, a numerically stable MLE training objective can be derived from Eq.~\ref{mle} as
\begin{align}
    \mathcal{L}_{s} = \frac{1}{N}\frac{1}{M}\sum_{i,t} \sqrt{2}\exp{\left(-\frac{1}{2}\boldsymbol{s}_i\right)}\left| \hat{\boldsymbol{y}}_{ti} - \hat{\boldsymbol{\mu}}_i \right| + \frac{1}{2}\boldsymbol{s}_i
\end{align}
where $i$ and $t$ corresponds to the summation over the output space and the generated samples respectively and $N$ and $M$ are number of instances (e.g pixels) in the output space and the generated predictive samples from teacher, respectively. The reason to choose Laplace distribution over Gaussian distribution is because it is more appropriate to model the variances of residuals with $\ell_1$ loss, which usually outperforms $\ell_2$ loss in computer vision tasks. 

For classification problems, we use a logit-normal distribution to model teacher's approximate predictive distribution $q(\boldsymbol{y} | \boldsymbol{x}, \mathcal{D_{\text{train}}})$ on the simplex\footnote{Dirichlet distribution is an obvious alternative, but we empirically observe that training with logit-normal is much more numerically stable.}. In practice, we use a Gaussian distribution with a diagonal covariance matrix to approximate the teacher's predictive distribution on the logit space. As a result, the student model outputs $\boldsymbol{\mu}_i$ and $\boldsymbol{s}_i$ as the mean and log variance of the Gaussian for each member of the logits. Similar to the regression set-up, we derive a numerically stable Gaussian MLE training objective
\begin{align}
    \mathcal{L}_{s} = \frac{1}{N}\frac{1}{M}\sum_{i, t} \frac{1}{2}\exp{\left(-\boldsymbol{s}_i\right)}\left|| \hat{\boldsymbol{y}}_{ti} - \hat{\boldsymbol{\mu}}_i \right||^2 + \frac{1}{2}\boldsymbol{s}_i
\end{align}
where $\boldsymbol{y}_{ti}$ are predicted logits sampled from teacher. Since close-form solution does not exist for the moments of a logit-normal distribution, Monte Carlo sampling on the logit space is performed at test time to obtain uncertainty estimates. This only incurs a tiny computational overhead during inference as it amounts to multiple forward passes of one layer of the student network (the softmax function). As shown in experiments, the student model still has a large advantage in inference time over its teacher in addition to better performance.

We empirically observe that training solely with the above loss functions sometimes leads to sub-optimal predictive performance. This may be due to the noisy signal provided by the generated samples. Thus we leverage ground truth labels in addition to predictive samples from the teacher to stabilize the training of the student model. We use the loss function for which the teacher model is trained in conjunction with the $\mathcal{L}_{s}$, leading to the total loss
\begin{align}
    \mathcal{L}_{total} = \mathcal{L}_{s} + \lambda \mathcal{L}_{t},
\end{align}
where the $\lambda$ is a hyper-parameter to be tuned and  $\mathcal{L}_{t}$ corresponds to the categorical cross entropy loss for classification tasks or L1 loss for regression tasks. We found that $\lambda = 1$ generally performs well for our experiments.

\subsubsection{Additional Augmentation} 
\label{section:augmentation}
When the same training dataset is used for both the teacher and the student training, the student may underestimate the epistemic uncertainty of the teacher due to overfitting of the teacher network to the training data. Ideally, in order to fully capture the teacher predictive distribution, the dataset used to train the student should not overlap with the one for the teacher. However, training using only a subset of available samples can lead to sub-optimal performance. To alleviate this problem, we perturb the training set during the training of the student using extra data augmentation methods unused when training the teacher, in order to synthetically generate new samples unseen by the teacher model. We choose color jittering as the augmentation method that augments each image via color jitter with random variation in the range of $[-0.2,0.2]$ in four aspects: brightness, contrast, hue, and saturation when training the student. 

As we demonstrate below, this extra augmentation during student training can be crucial for enhanced quality of uncertainty estimates. 
We emphasize that the additional gain in uncertainty estimates does not directly come from data augmentation, but rather from teacher predictions that more closely correspond to the test-time predictive distributions as a consequence of this augmentation. 
In the experiments below, we show that the teacher model does not have the same performance boost with the extra augmentation.
\begin{figure}[t!]
\centering
\includegraphics[width=1\linewidth]{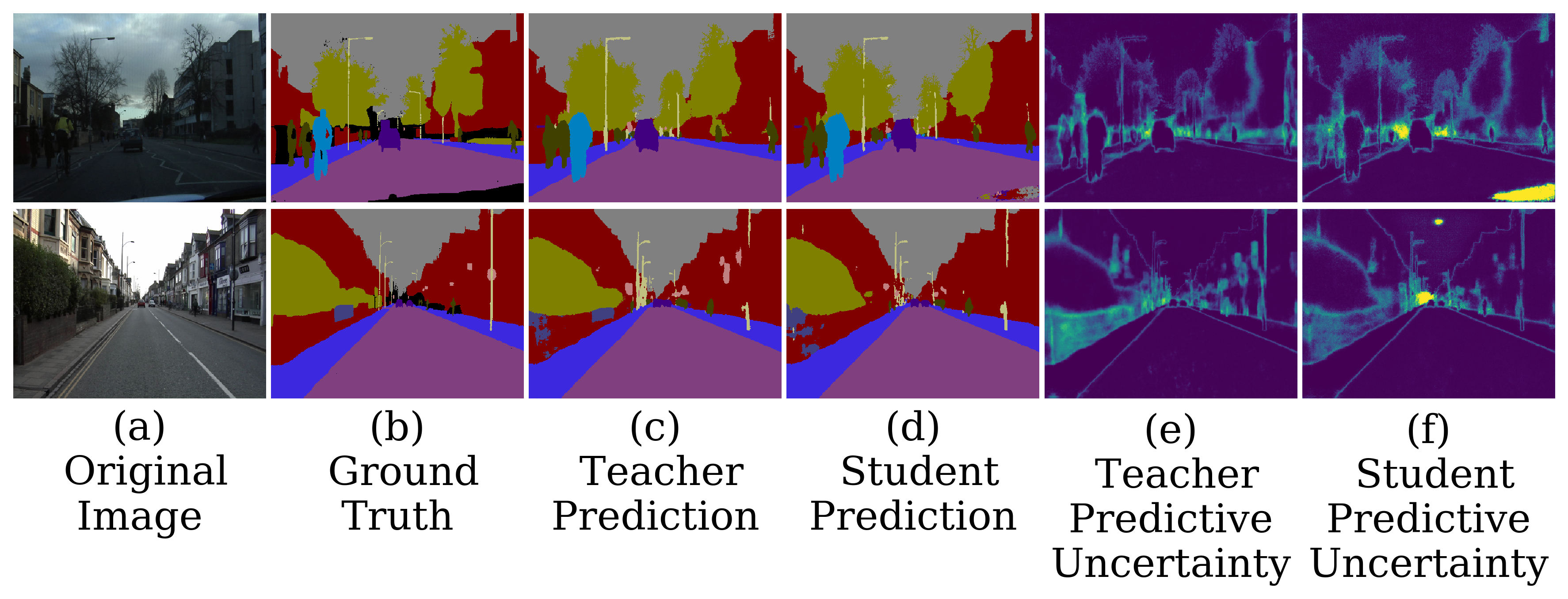}
\caption{Example predictions on CamVid. Each uncertainty map shows the sum of aleatoric and epistemic uncertainty. Same for all the following example plots.}
\label{camvid_figure}
\end{figure}
\section{Experiments}
We conduct experiments on two pixel-wise computer vision tasks: semantic segmentation and depth regression. We compare the performance of the proposed method with that of the teacher models using MC dropout. For a holistic evaluation, we consider teacher networks trained both with and without the aleatoric uncertainty. Following\cite{kendall2017uncertainties, postels2019sampling}, we use 50 samples for MC dropout to evaluate teacher's performance and uncertainty. Architectures identical to that of the teacher models without the dropout layers are used as student models. As discussed in the previous section, we use 50 samples from the logit space to evaluate uncertainty (BALD) of student models for classification tasks. To demonstrate the general applicability, we also show the effectiveness of the proposed method when the teacher network corresponds to a Deep Ensemble\cite{lakshminarayanan2017simple}. 

\subsubsection{Evaluation Metrics} 
On top of metrics to evaluate the performance of the predictive means of our models, we measure both the Area Under the Sparsification Error curve (AUSE)\cite{ilg2018uncertainty} and the expected calibration error (ECE)\cite{guo2017calibration} as measures to evaluate the quality of uncertainty estimates. In essence, AUSE measures how much the estimated uncertainty coincides with true predictive errors. Brier score and the mean absolute error are used as predictive errors to compute AUSE for classification and regression tasks respectively. In the context of classification, ECE measures how much the predictive means of probabilities from the softmax function are representative of the true correctness of predictions. In the context of regression, we use ECE described in\cite{kuleshov2018accurate} to quantify the amount of mismatch between the predictive distribution and the empirical CDFs. We follow\cite{kuleshov2018accurate} and compute ECE with the $\ell_2$ norm with a bin size of 30. 

\subsection{Semantic Segmentation}
Bayesian SegNet\cite{kendall2015bayesian}, which contains dropout layers inserted after the central four encoder and decoder units, was proposed to obtain uncertainty estimates for semantic segmentation. In this work, we use the architecture with a dropout rate of $0.5$ as the teacher model for all our experiments. We use the CamVid and VOC2012 datasets. 

For CamVid, following Kendall et al.\cite{kendall2015bayesian}, we use $11$ generalized classes and a downsampled image size of $360 \times 480$. For the teacher network, we train using the Stochastic Gradient Descent (SGD) with an initial learning rate of $10^{-3}$, a momentum of $0.9$, and a weight decay of $5 \times 10^{-4}$ for 100,000 steps. In order to achieve faster convergence, we initialize the student network using the weights of the teacher network. To this end, a smaller initial learning rate of $5\times10^{-4}$ is used to train the student network for 80,000 steps. We employ a ``poly" learning rate policy on both the teacher and student networks as done by Chen et al.\cite{chen2017rethinking}. We use a batch size of 4 for both per step. 

For VOC2012, we use the same augmented ``train" and ``val" split as in\cite{chen2017rethinking}. Input images are resized to $224 \times 224$. For the optimal performance of the teacher model, SGD with a higher initial learning rate of $10^{-2}$ is used instead, with a batch size of 8 for $150000$ steps. Similarly, we initialize the student model with the weights of the teacher. The student model is trained for $100000$ steps with an initial learning rate of $10^{-3}$ using a size of 8 per step. The performance on the ``val" split is reported in the results.
We also include the results of the student models trained using Dropout Distillation (DD) \cite{bulo2016dropout} as a baseline comparison.
\begin{figure}[t!]
\centering
\includegraphics[width=1\linewidth]{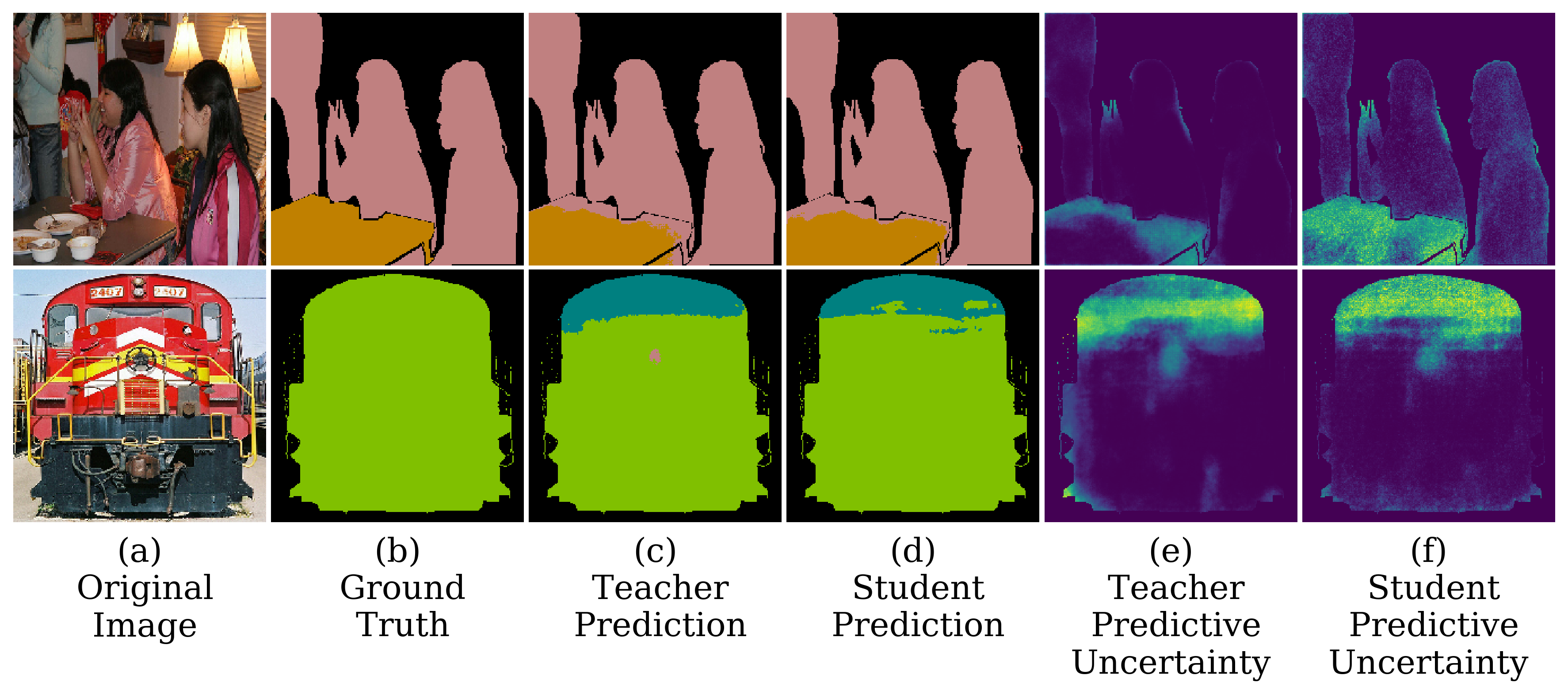}
\caption{Example predictions on Pascal VOC2012.}
\label{pascal_figure}
\end{figure}

\subsubsection{Evaluation} 
Results for both the teacher and the student are summarized in Table~\ref{camvid_table}. On top of a significant boost in run-time, the student network also leads to improvements in terms of most of the metrics evaluated. We believe the reason for the observed improvements in both predictive performance and uncertainty estimates is mainly due to learning the entire predictive distribution implicitly through samples from the teacher models with the proposed optimization objective can have the \textit{loss attenuation} effect as described in\cite{kendall2017uncertainties}. In contrast, Dropout Distribution (DD) \cite{bulo2016dropout}, which only distills the mean prediction of the teacher as the standard knowledge distillation, shows worse performances of the student than those of the teachers in all the metrics. This further demonstrates the benefit of distilling the entire predictive distribution from the teacher.
\begin{table}[t!]
\centering
\caption{Results on the segmentation problem. The ``T'', ``S'' and ``AU'' corresponds to the teacher and student model, and the aleatoric uncertainty respectively. ``T+AU'' corresponds to a teacher model trained with the aleatoric uncertainty. ``DD'' corresponds to the student trained using Dropout Distillation \cite{bulo2016dropout}. Best performing results for each teacher-student pair are bold-faced.}
\begin{adjustbox}{width=1.0\textwidth}
\begin{tabular}{c|ccc|cc}
\hline 
   \multicolumn{6}{c}{\textbf{Camvid}} \\ \hline \hline
Model         & T   & S  &DD\cite{bulo2016dropout} & T+AU   & S+AU\\ \hline
Accuracy $\uparrow$ & 0.906 &  \textbf{0.907}  & 0.903 & 0.907  & \textbf{0.909}       \\ 
Classwise Acc $\uparrow$ & 0.764    &  \textbf{0.765}   & 0.747 &   \textbf{0.766}     & 0.750 \\ 
IOU $\uparrow$ &  0.645   &  \textbf{0.650}  & 0.642 &   0.645     & \textbf{0.650}  \\ 
ECE $\downarrow$ \tiny{($\times 10^{-3}$)}&  3.78   &  \textbf{2.23}  & 6.73 &    3.67    &   \textbf{2.86}   \\
AUSE $\downarrow$ \tiny{($\times 10^{-2}$)}   &  \textbf{1.47}  &  1.60  & 2.59 &  1.63  &  \textbf{1.60}  \\ 
Runtime (s) $\downarrow$  &  1.6   &  \textbf{0.078} & \textbf{0.078}   &   2.1     &  \textbf{0.078}  \\ \hline 
\end{tabular}
\end{adjustbox}
\begin{adjustbox}{width=1.0\textwidth}
\begin{tabular}{c|ccc|cc}
\hline 
 \multicolumn{6}{c}{\textbf{Pascal VOC}} \\ \hline \hline
Model         & T   & S  &DD\cite{bulo2016dropout} & T+AU   & S+AU \\ \hline
Accuracy $\uparrow$ & 0.834    &   \textbf{0.851}    &0.828  &    0.831      & \textbf{0.848}       \\ 
Classwise Acc $\uparrow$ & 0.813 & \textbf{0.828} & 0.806  & 0.809 & \textbf{0.827}  \\ 
IOU $\uparrow$ & 0.697  & \textbf{0.727} & 0.691 & 0.693  &  \textbf{0.722}   \\ 
ECE $\downarrow$ \tiny{($\times 10^{-3}$)}&  62.7    &  \textbf{59.0}   & 67.5  &  63.0        &  \textbf{59.0}    \\
AUSE $\downarrow$ \tiny{($\times 10^{-2}$)} & 4.35 & \textbf{3.82}  &4.86 & \textbf{4.20}  & 4.31    \\ 
Runtime (s) $\downarrow$  &   0.51   &   \textbf{0.028} &   \textbf{0.028}   &   0.68       & \textbf{0.028}   \\ \hline 
\end{tabular}
\end{adjustbox}
\label{camvid_table}
\end{table}

Figure \ref{camvid_figure} and \ref{pascal_figure} are random selected examples from the validation set of CamVid and Pascal VOC respectively. Visual examples suggest that the student model can accurately capture both the predictive mean and uncertainty of the teacher model. Furthermore, a closer comparison reveals the exceptional quality of the uncertainty estimates produced by the student model. For instance, in the second example from the CamVid dataset in Figure \ref{camvid_figure}, a small part of the ego vehicle is captured by the camera at the bottom of the figure. While the teacher model confidently predicts the area as ``road surface'', the student model highlights this subtle anomaly with high uncertainty estimates. A similar contrast is also observed in the top example of Figure \ref{pascal_figure}, where the boundary of people is assigned much higher uncertainty by the student model. Besides, the bowls and plates on the dining table, which are not in the list of labeled classes for the dataset, also ``confuses" the student model, but not the teacher.

\subsubsection{Run-Time Comparison} 

\begin{figure}
\centering
\begin{subfigure}{.45\textwidth}
\centering
\includegraphics[width=1\linewidth]{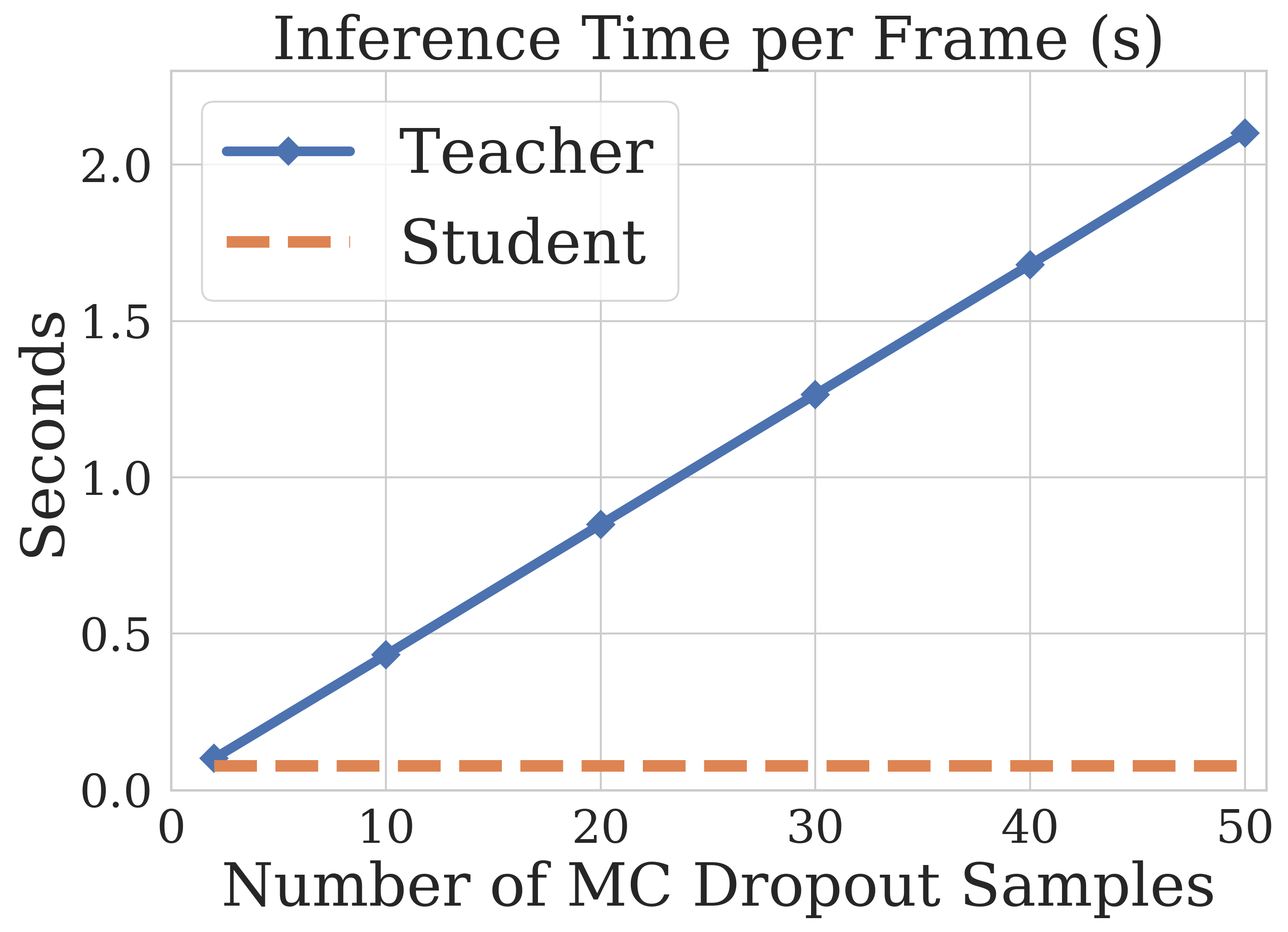}
\caption{}
\end{subfigure}
\begin{subfigure}{.45\textwidth}
\centering
\includegraphics[width=1\linewidth]{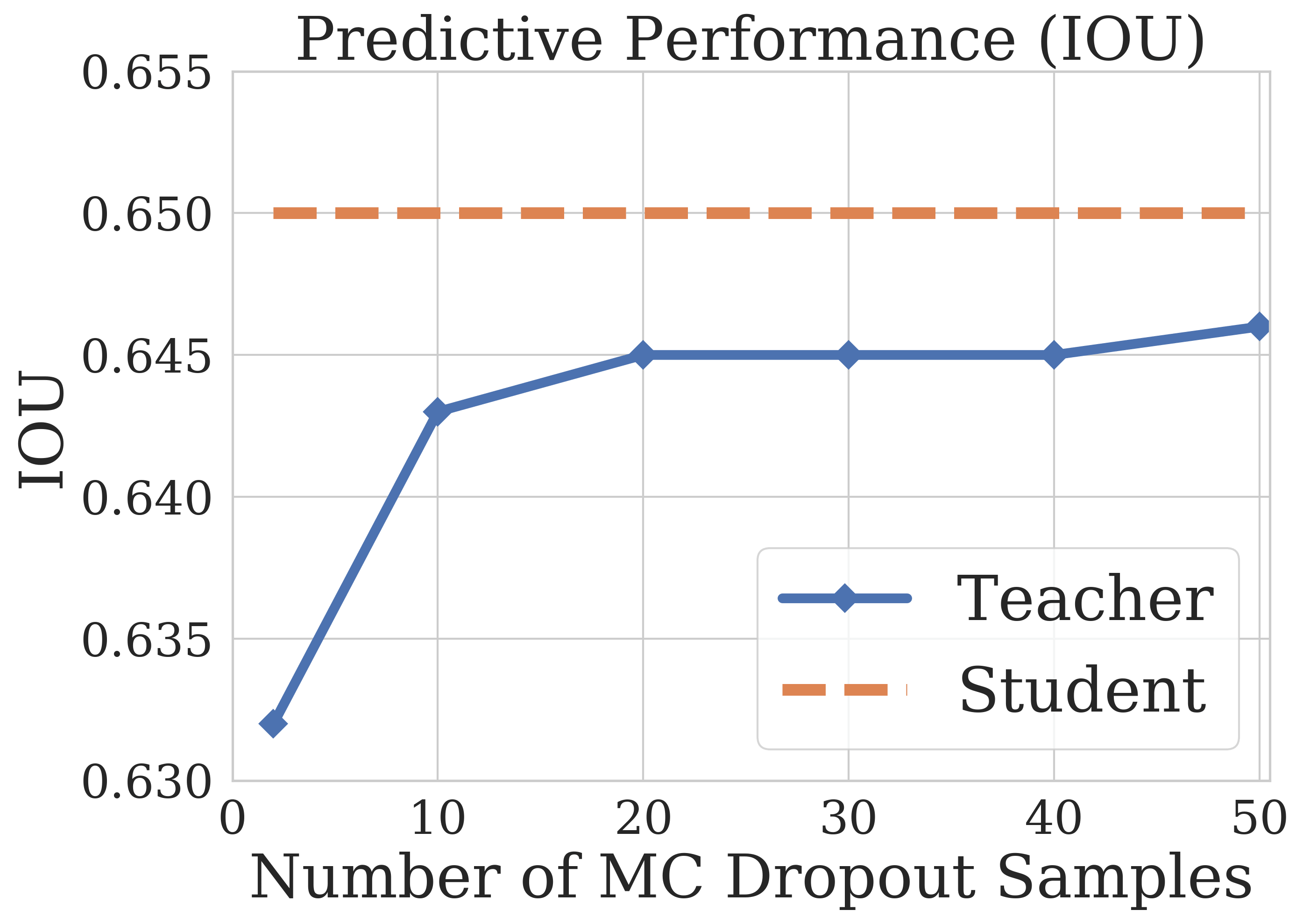}
\caption{}
\end{subfigure}
\begin{subfigure}{.45\textwidth}
\centering
\includegraphics[width=1\linewidth]{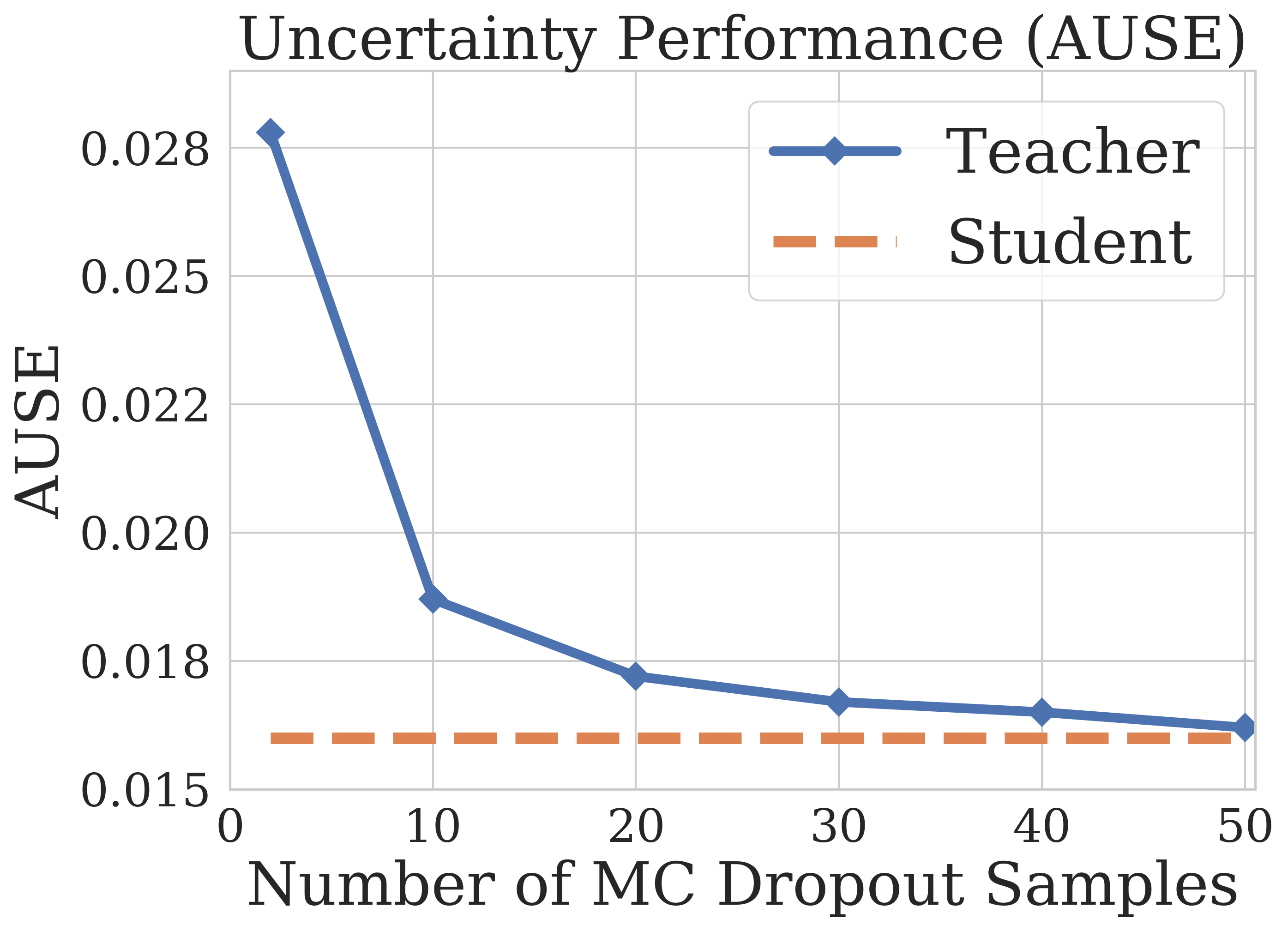}
\caption{}
\end{subfigure}
\begin{subfigure}{.45\textwidth}
\centering
\includegraphics[width=1\linewidth]{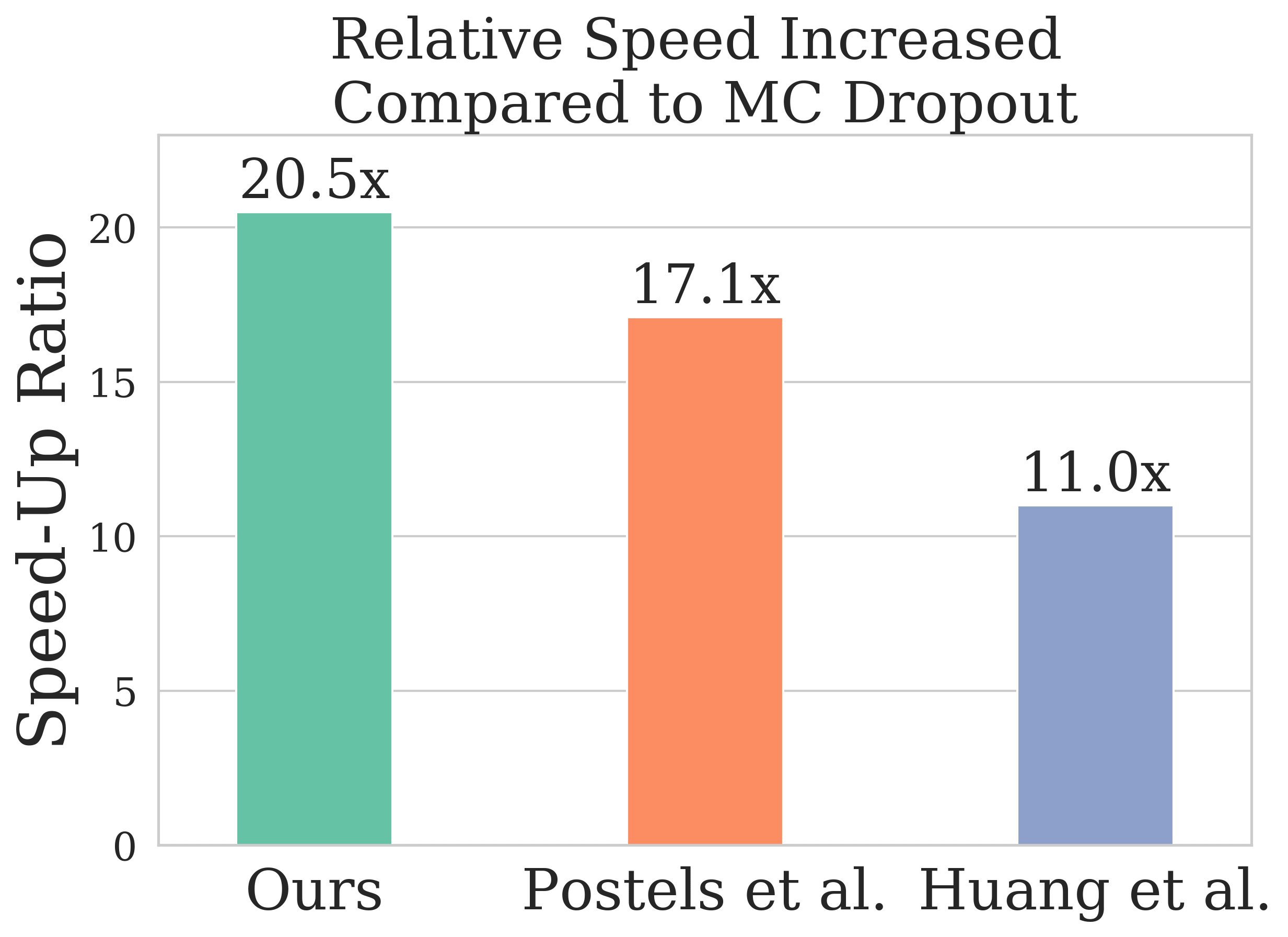}
\caption{}
\end{subfigure}
\caption{(a)-(c): Comparison of performance against the running time for both the teacher (with the aleatoric uncertainty) and student model using the CamVid dataset. (d) Speed-up ratios of uncertainty estimates for the CamVid dataset with the Bayesian SegNet compared to Huang et al.\cite{huang2018efficient} and Postels et al.\cite{postels2019sampling}, two other sample-free uncertainty estimation methods.}
\label{runtime_plot}
\end{figure}

Figure~\ref{runtime_plot} (a)-(c) illustrate a comparison of running time and performance using different numbers of samples for MC dropout. While the running time of MC dropout can be shortened with fewer samples, it comes at the cost of quality of prediction and uncertainty estimates. The running time of MC Dropout is optimized by caching results before the first dropout layer for a fair comparison. 

We further demonstrate the merit of the proposed method by comparing the running time of the student with several other recently proposed sample-free methods for uncertainty estimates. Figure~\ref{runtime_plot} (d) illustrates the speed boost with different methods on the CamVid dataset with Bayesian SegNet. The ratios are computed with respect to the same baseline of MC dropout with 50 samples at test time. Our proposed method achieves a more significant boost in speed than previously proposed methods for accelerating dropout inference, in addition to other advantages such as wider applicability and improved predictive performance.

\subsubsection{Performance under Distribution Shift}
\begin{figure}[t!]
\centering
\includegraphics[width=0.8\linewidth]{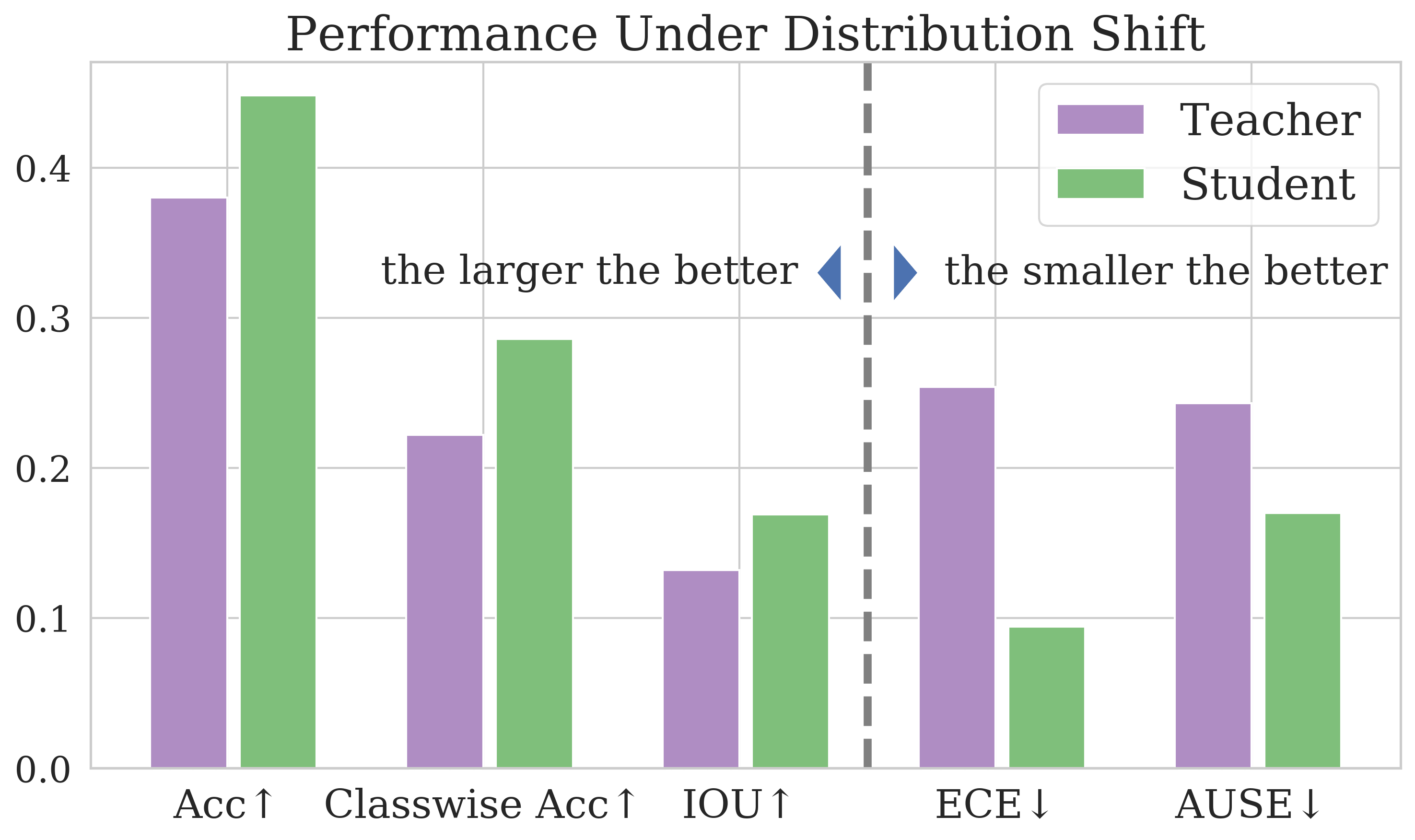}
\caption{Performance of models trained with CamVid and evaluated on Cityscapes.}
\label{distribution_shift}
\end{figure}
We also evaluate the performance of the proposed method under a distribution shift using models trained with the CamVid dataset. The Cityscapes dataset\cite{Cordts2016Cityscapes}, which contains street scenes collected from different cities, is an ideal dataset for such evaluation. We emphasize that neither the teacher nor the student sees images from the Cityscapes dataset during training. The results are summarized in Figure \ref{distribution_shift}, which is evaluated on the overlapped classes between CamVid and Cityscapes. Surprisingly, while both the teacher and student models perform unsatisfactorily, the student performs significantly better than the teacher in terms of all of the metrics evaluated, suggesting its enhanced robustness against the distribution shift when trained with the proposed teacher-student pipeline. We hypothesis that by seeing the distribution of soft labels from a bayesian teacher from the distillation process, the student learns to output less confident, more generalizable outputs. The true cause can leave for further works. This can be important for lots of application domains with long-tail scenarios like autonomous driving. 

\subsubsection{Outlier Detection}
\begin{figure}[t!]
\centering
\begin{subfigure}{.45\textwidth}
\centering
\includegraphics[width=1\linewidth]{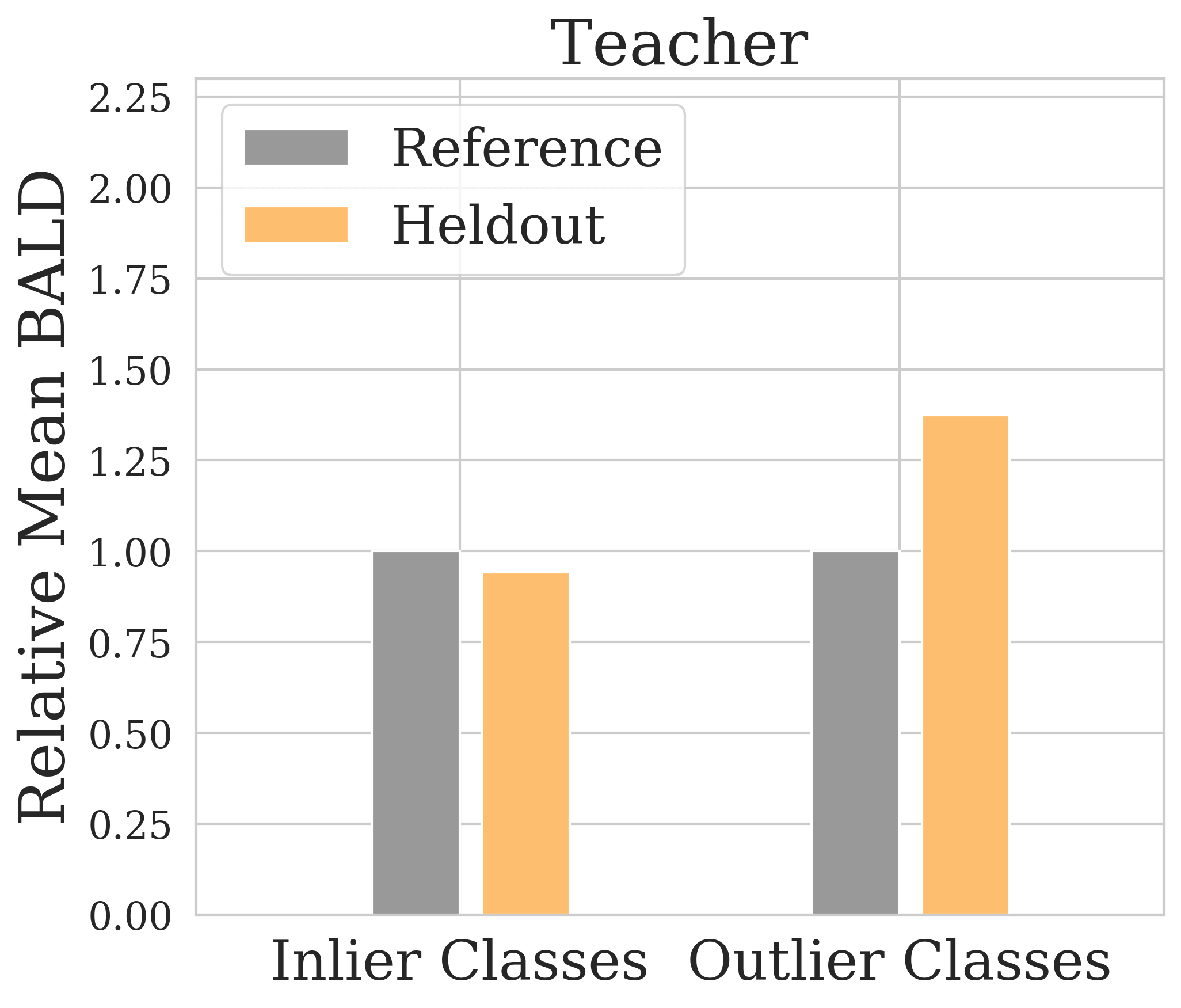}
\end{subfigure}
\begin{subfigure}{.45\textwidth}
\centering
\includegraphics[width=1\linewidth]{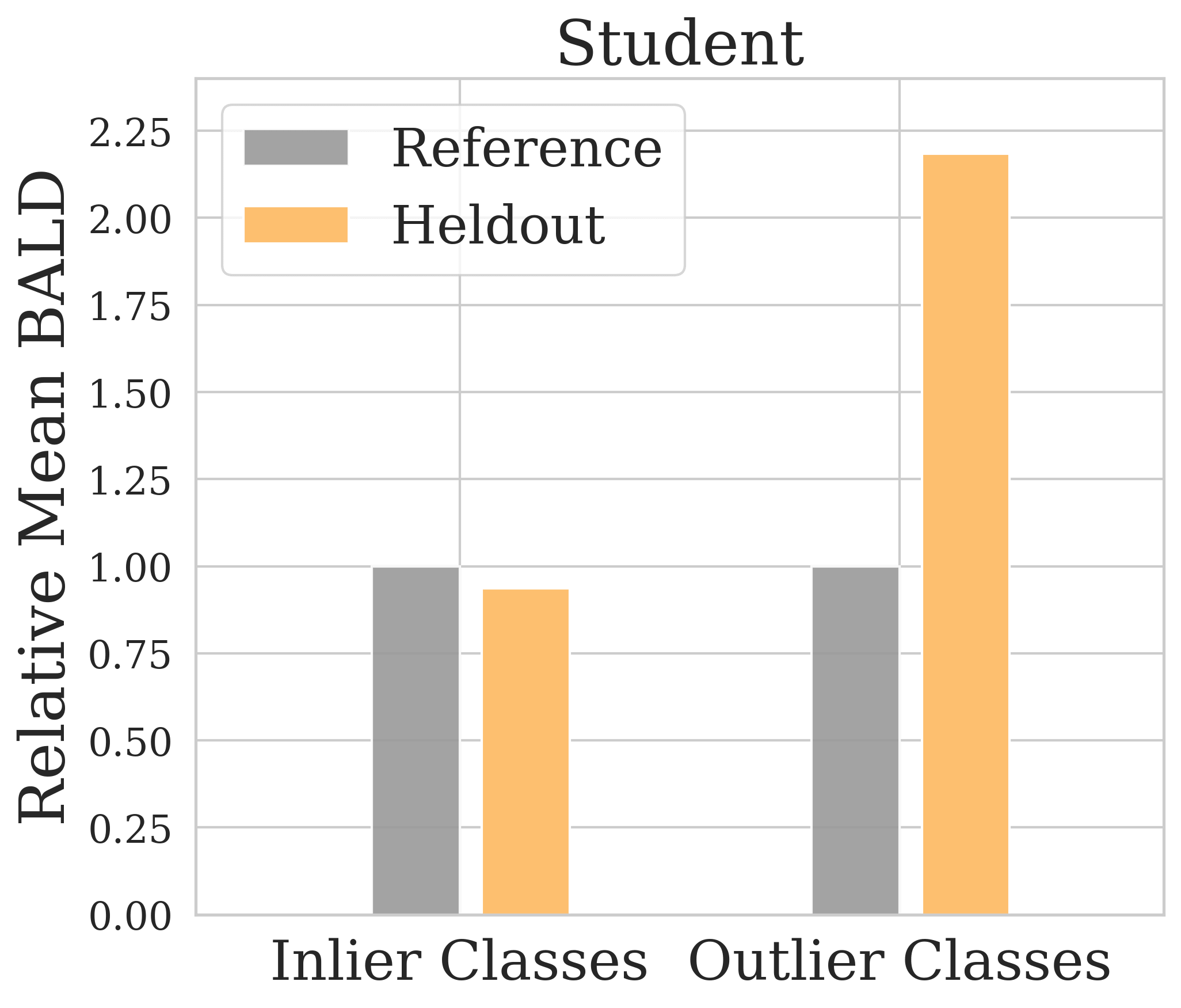}
\end{subfigure}
\begin{subfigure}{.45\textwidth}
\centering
\includegraphics[width=1\linewidth]{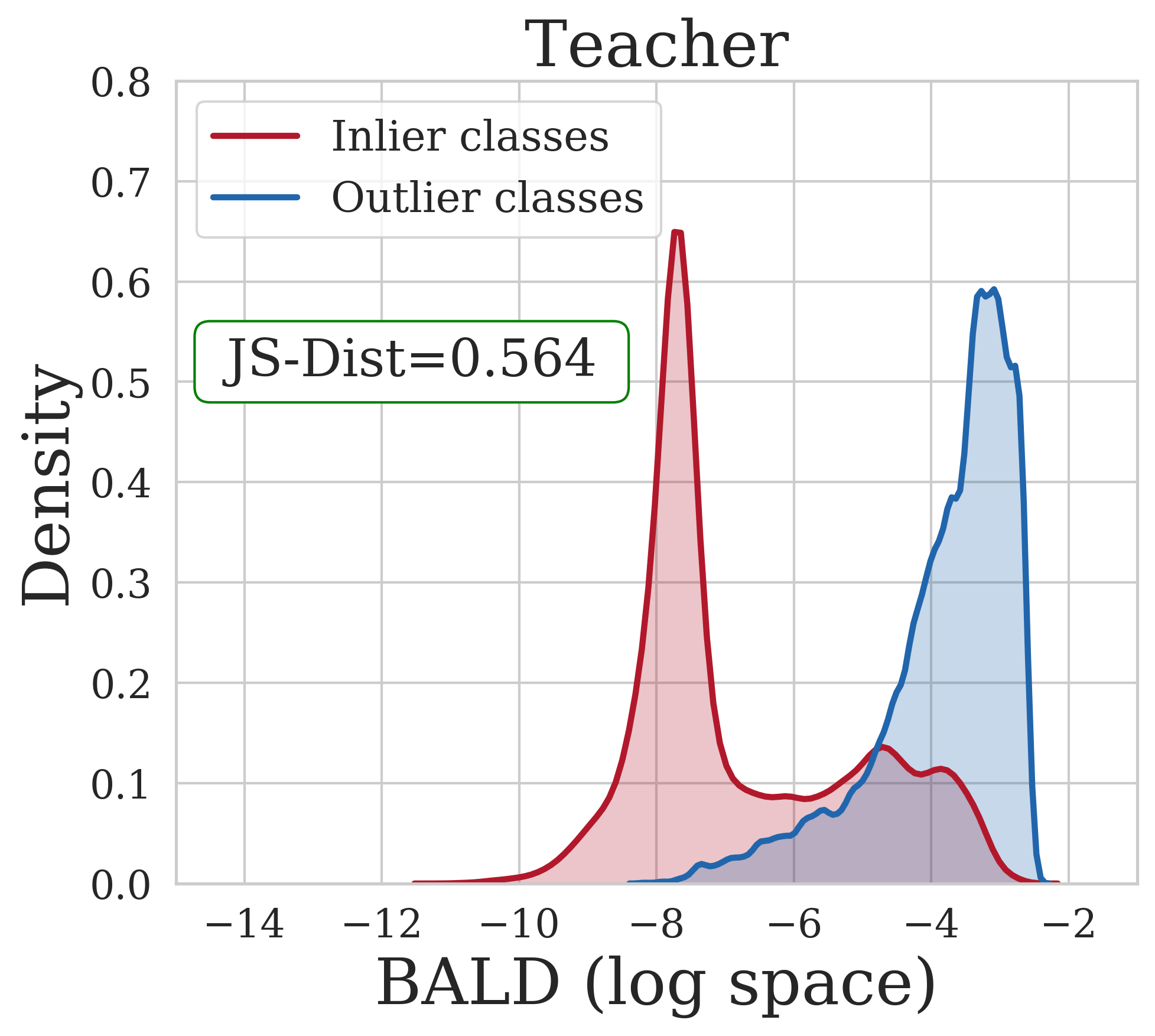}
\end{subfigure}
\begin{subfigure}{.45\textwidth}
\centering
\includegraphics[width=1\linewidth]{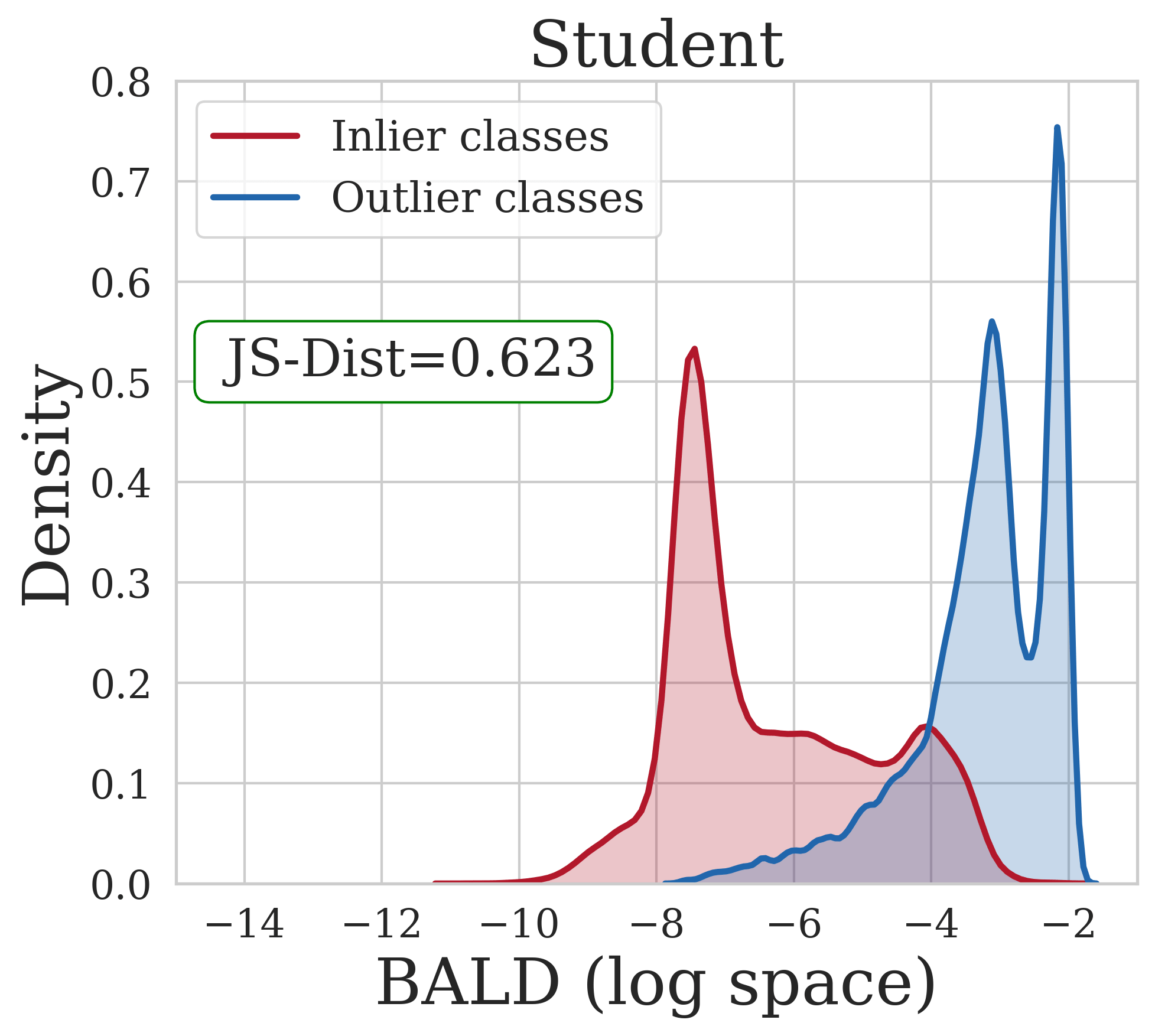}
\end{subfigure}
\caption{\textit{Top}: Relative means of BALD for samples of seen and unseen classes during training compared to the ``Reference'' models, which refer to models trained with both seen and unseen classes. \textit{Bottom}: Distribution of BALD for samples of seen and unseen classes during training.}
\label{ood_plot}
\end{figure}
\begin{figure}[t!]
\centering
\includegraphics[width=1\linewidth]{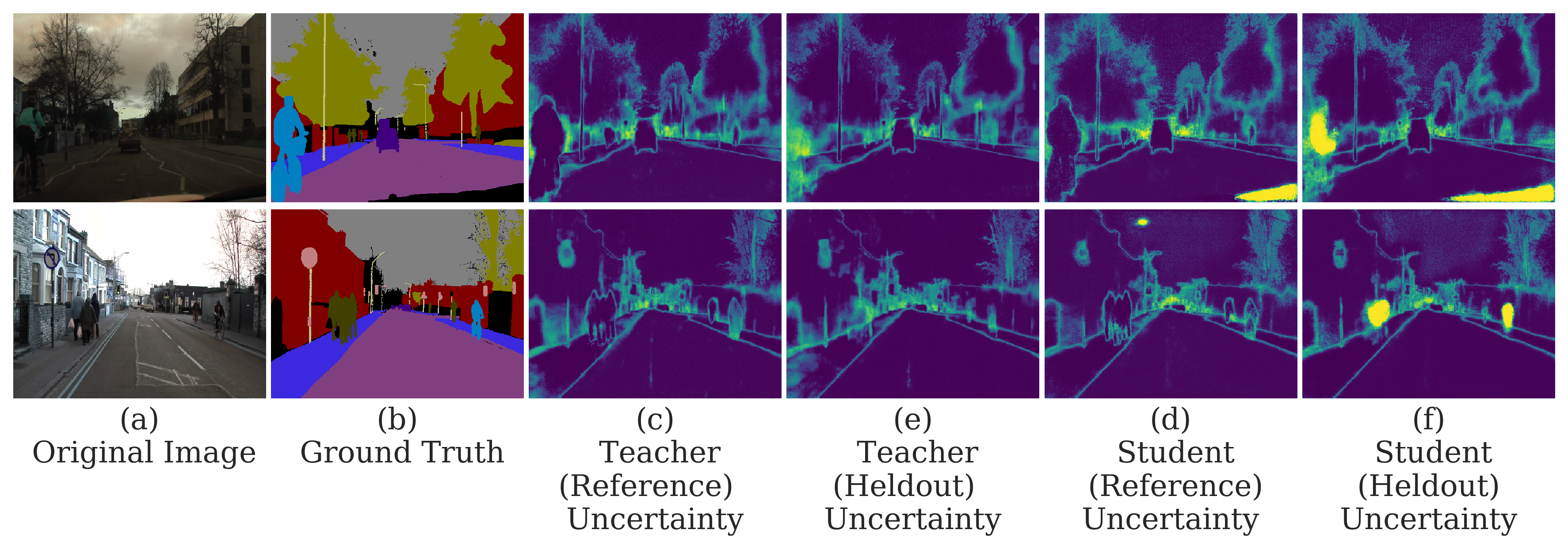}
\caption{Example predictions on CamVid when ``pedestrian'' and ``bicyclist'' are held out during training. ``Reference'' refers to models trained with all classes.}
\label{ood_example}
\end{figure}
In addition, we examine the effectiveness of the uncertainty estimates for outlier detection using the CamVid dataset. Following\cite{postels2019sampling}, we use ``pedestrian'' and ``bicyclist'' as held-out classes and exclude them from training. Ideally, classes unseen during training should have much higher uncertainty estimates than that of the seen classes. We show in Figure \ref{ood_plot} comparisons of relative means of the uncertainty estimates against those of ``reference'' models, which refer to models trained with both seen and unseen classes, for both inlier and outlier classes. While both teacher and student assign higher uncertainty to outlier classes compared to the ``reference'' models on average, the relative mean is much higher for the student. To further quantify the performance, we also compute the Jensen–Shannon distance between distributions of uncertainty estimates of inlier and outlier classes\cite{maddox2019simple}. 
Again, the difference in the inlier and outlier distribution is larger for the student network, suggesting its enhanced ability for outlier detection. Lastly, we show in Figure \ref{ood_example} two randomly chosen examples to illustrate the difference between teacher and student. As seen clearly, regions with pedestrians and bicyclists have higher uncertainty estimates when they are not present in training for both the teacher and student. The magnitude is much larger for the student as represented by bright spots in the uncertainty plot.

\subsection{Pixel-Wise Depth Estimation}
For pixel-wise depth estimation tasks, NYU DEPTH V2 (NYU) and the KITTI Odometry dataset (KITTI) are used to conduct experiments. We follow the same ResNet-based architecture to\cite{mal2018sparse} for the training of both datasets in RGB based depth estimation, with dropout $p =
0.2$ placed after each convolutional layer except the final one. For NYU, we use the same train/test split as in\cite{mal2018sparse} and for KITTI we train our models on sequences 00-10 and evaluate them on sequences 11-21. Identical procedures are used to train the teacher models for both NYU and KITTI. During training, SGD optimizer with an initial learning rate of $0.01$, a momentum of $0.9$, and weight decay of $10^{-4}$ with "poly" learning rate poly is adopted for a total of 40 epochs. For NYU, we initialize the student model with the weights of the teacher and train the student model for 30 epochs using a smaller learning rate of $0.005$. We empirically observe that initializing with the teacher model for KITTI leads to overfitting to the training set and thus we train the student model from scratch with the identical procedure as used for teacher training. We use a batch size of 8 in all of the depth estimation experiments.

\subsubsection{Evaluation}
\begin{table}[t!]
\centering
\caption{Results on the depth estimation. 
The ``T'', ``S'' and ``AU'' corresponds to the teacher and student model, and the aleatoric uncertainty respectively. ``T+AU'' corresponds to a teacher model trained with the aleatoric uncertainty. 
}
\begin{adjustbox}{width=1.0\textwidth}
\begin{tabular}{l|cc|cc|cc|cc}
\hline 
              & \multicolumn{4}{c|}{\textbf{NYU}} & \multicolumn{4}{c}{\textbf{KITTI}} \\ \hline \hline
Model         & T   & S   & T+AU   & S+AU  & T     & S     & T+AU     & S+AU     \\ \hline
RMSE $\downarrow$ & 0.542 &  \textbf{0.540}  &   \textbf{0.548}     &  \textbf{0.548}     &   4.80    &   \textbf{4.75}    &    4.83      & \textbf{4.81}          \\ 
REL $\downarrow$ & 0.155  & \textbf{0.152} & 0.158 & \textbf{0.154} & 0.123 & \textbf{0.122} & \textbf{0.117} & \textbf{0.117}   \\ 
$\log 10$ $\downarrow$ & 0.065 & \textbf{0.064} & 0.065 & \textbf{0.064} & 0.053 & \textbf{0.052} & 0.052 & \textbf{0.051}  \\ 
$\delta 1$ $\uparrow$ & 0.793 & \textbf{0.798} & 0.794 & \textbf{0.799} & 0.843 & \textbf{0.847} & 0.845 & \textbf{0.846}     \\ 
$\delta 2$  $\uparrow$   & 0.947 & \textbf{0.949} & 0.945 & \textbf{0.946} & 0.948 & \textbf{0.951} & \textbf{0.950} & 0.949   \\ 
$\delta 3$ $\uparrow$  & \textbf{0.985} & 0.984 & \textbf{0.982} & 0.981 & 0.981 & \textbf{0.982} & \textbf{0.981} & \textbf{0.981}   \\ 
ECE $\downarrow$ \tiny{($\times 10^{-2}$)} & 9.38  & \textbf{8.09}  & 5.79  & \textbf{5.13}  & 7.80  & \textbf{2.95}  & 4.53  & \textbf{2.18}  \\
AUSE $\downarrow$ \tiny{($\times 10^{-2}$)} & \textbf{6.01}  &6.06  & 5.88  & \textbf{5.82}  &  0.701  & \textbf{0.660}  & 0.597  & \textbf{0.595}  \\ 
Runtime (s)  $\downarrow$ &  0.73  & \textbf{0.016}  & 0.739  & \textbf{0.016}  &  0.28  & \textbf{0.007}  & 0.29 & \textbf{0.007}  \\ \hline 
\end{tabular}
\end{adjustbox}
\label{depth_table}
\end{table}

\begin{table}[t!]
\centering
\caption{\textit{Top-4 Rows}: Impact of adding augmentation in training on quality of uncertainty produced on the CamVid and NYU datasets. "T" and "S" represents teacher and student models, and "AUG" corresponds to augmentation. \textit{Last Row}: Uncertainty performance of student model when a deep ensemble with five NNs is used as the teacher model.}
\begin{adjustbox}{width=1.0\textwidth}
\begin{tabular}{l|cc|cc}
\hline
  & \multicolumn{2}{c|}{\textbf{CamVid}}  & \multicolumn{2}{c}{\textbf{NYU}} \\ \hline \hline 
   & ECE \tiny{($\times 10^{-3}$)}  & AUSE \tiny{($\times 10^{-2}$)}  & ECE \tiny{($\times 10^{-3}$)}   & AUSE \tiny{($\times 10^{-2}$)}  \\ \hline
T w/o AUG & 3.67 & 1.62  &   57.9    &   5.88   \\
T w/ AUG & 3.90 & 1.62  &   57.1    &   5.90    \\ \hline 
S w/o AUG  & 4.63 & 2.19 &  54.0 & 5.91    \\
S w/ AUG  & \textbf{2.86} &\textbf{1.60} &  \textbf{51.3}  &  \textbf{5.80}    \\ \hline \hline 
S w/ Ens T & 2.96 & 1.91 & 56.3 & 5.93   \\ \hline
\end{tabular}
\end{adjustbox}
\label{augmentation_table}
\end{table}
\begin{figure}[t!]
\centering
\includegraphics[width=1\linewidth]{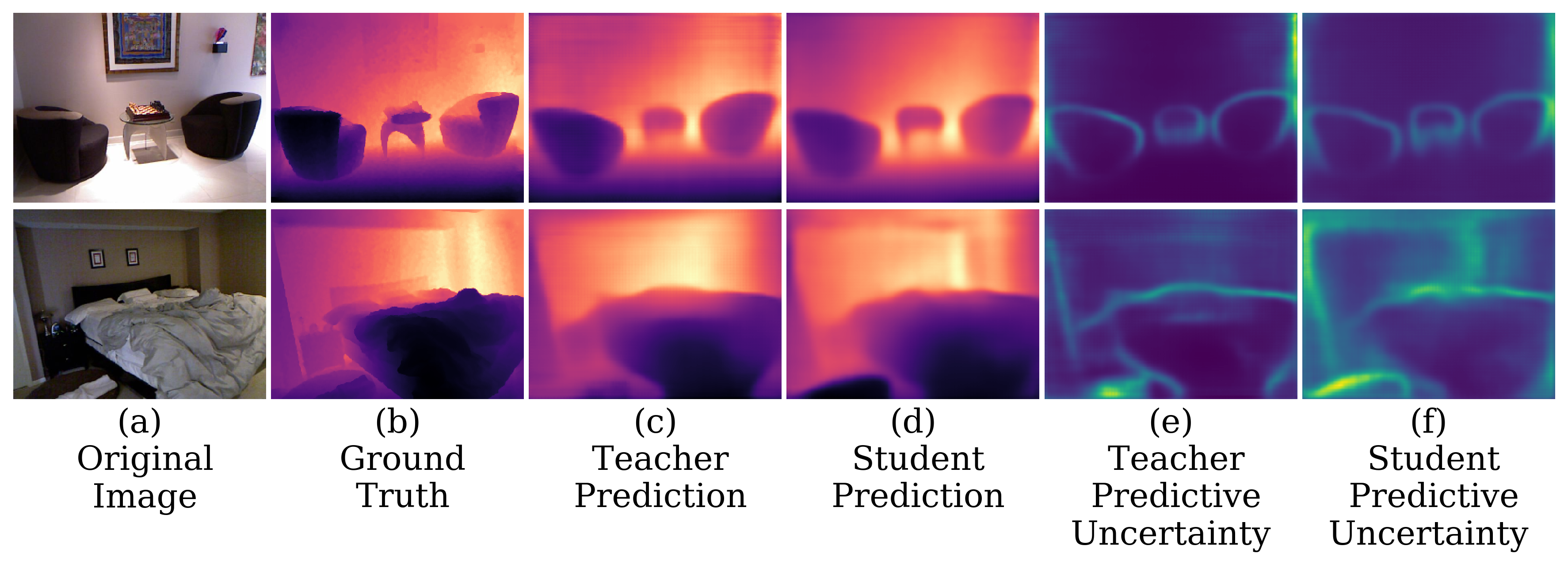}
\caption{Example predictions on NYU.}
\label{nyu_figure}
\end{figure}
The quantitative performance of both the teacher and the student models is summarized in Table \ref{depth_table}. Similar to segmentation tasks, the student model outperforms the teacher in most of the evaluation metrics.  
Example predictions shown in Figure \ref{nyu_figure} again illustrate that the student network is able to closely approximate the uncertainty estimates produced by the teacher model.
Moreover, as more number of dropout layers are inserted into the NNs for experiments with depth estimation, the relative speed-up ratio achieved by the student model is further increased due to less cached computation for the teacher. For instance, the student model achieves a speed-up ratio of $46$ for the NYU dataset. 

\subsection{Ablation Study on Additional Augmentation}
To demonstrate the importance of additional augmentation during student training, we also summarize in Table \ref{augmentation_table} results when the student is trained without extra augmentation. Using extra augmentation in the student training process as discussed in Section~\ref{section:augmentation} helps the student produce much better uncertainty estimation. We can also see that the same extra augmentation does not improve the performance of the teacher's uncertainty estimation, suggesting that the student model benefits from seeing the teacher's predictions more closely aligned with the test-time predictive distributions, rather from data augmentation itself.

\subsection{Distilling from Deep Ensemble}
To examine the effectiveness of using deep ensembles as teachers\cite{malinin2019ensemble}, we train an ensemble of deterministic neural networks with aleatoric uncertainty\cite{lakshminarayanan2017simple}. The training detail is identical to that described above. Due to limited computational resources, we fix the number of models in the ensemble to five. Dirichlet distribution is not used to approximate teacher's predictive distribution for classification as in \cite{malinin2019ensemble} because we empirically found it very numerically unstable and led to failure of convergence. We show the uncertainty results in Table 3. Full results can be found in the Appendix. As seen clearly from Table 3, the student obtained from the ensemble teacher have worse calibration performance than the student distilled from MC-Dropout teachers. The gap is likely due to the difficulty in learning a good predictive distribution with just 5 samples.

\subsection{Discussion}
Our experiments show that incorporating aleotoric uncertainty can result in minimal improvements for both teachers and students. This could be caused by significant overlaps between the two types of uncertainties learned by the teacher model, since the two types of uncertainties are not mutually exclusive, and can coincide significantly\cite{kendall2017uncertainties}. 
Nonetheless, aleatoric uncertainty can be beneficial for other tasks and datasets. The goal of the paper is to propose a general distillation strategy capable of also incorporating aleotoric uncertainty. Using the proposed approach, as clearly seen from the experimental results, students can match or surpass their teacher models in performance with or without aleatoric uncertainty. 

We also stress that, since the student is supervised by both the ground truth labels and the teacher's predictions, there can be discrepancies in predictive distributions between the teacher and the student models. Nevertheless, we believe these discrepancies can be beneficial and account for the improved performance of the student. As demonstrated, student models produce well-calibrated uncertainty maps that also semantically make sense without the need for expensive multiple forward passes.

\section{Conclusion}
We presented a two-stage teacher-student framework for fast uncertainty estimates. The proposed student training procedure is not only capable of producing uncertainty estimates at no extra cost but also leads to improved predictive performance and more informative uncertainty estimates. We believe the method gets us one step closer to the realm of trustworthy deep learning for computer vision.

%
%
{\small
\bibliographystyle{ieee_fullname}
\bibliography{egbib}
}

\clearpage
\onecolumn

\end{document}